\documentclass[letterpaper, 10 pt, conference]{ieeeconf}  
\IEEEoverridecommandlockouts                              
\usepackage{comment}
\usepackage{xcolor}

\title{\LARGE \bf
Exploring Collaborative Game Play with Robots to Encourage Good Hand Hygiene Practises among Children}

\author{Devasena Pasupuleti$^{1}$, Sreejith Sasidharan$^{2}$, Rajesh Sharma$^{3}$ and Gayathri Manikutty$^{4}$ 
\thanks{$^{1}$Devasena Pasupuleti, $^{2}$Sreejith Sasidharan, $^{4}$Gayathri Manikutty are with AMMACHI Labs, Amrita Vishwa Vidyapeetham, Kollam, Kerala, 690525, India}
\thanks{{\tt\small $^{1}$pdevasena835@gmail.com}}
\thanks{{\tt\small $^{2}$ssreejith172@gmail.com}}
\thanks{{\tt\small $^{4}$gayathri.manikutty@ammachilabs.org}}
\thanks{$^{3}$Rajesh Sharma is with Spire Animation Studios, Los Angeles, California, 91403, United States of America}
\thanks{{\tt\small $^{3}$rajesh.k.sharma@gmail.com}}
}

\begin{document}
\maketitle
\thispagestyle{empty}
\pagestyle{empty}

\begin{abstract}
This paper presents the design, implementation, and evaluation of a novel collaborative educational game titled ``Land of Hands", involving children and a customized social robot that we designed (\emph{HakshE}). Through this gaming platform, we aim to teach proper hand hygiene practises to children and explore the extent of interactions that take place between a pro-social robot and children in such a setting. We blended gamification with Computers as Social Actors (CASA) paradigm to model the robot as a social actor or a fellow player in the game. The game was developed using Godot's 2D engine and Alice 3. In this study, 32 participants played the game online through a video teleconferencing platform \emph{Zoom}. To understand the influence a pro-social robot's nudges has on children's interactions, we split our study into two conditions: With-Nudges and Without-Nudges. Detailed analysis of rubrics and video analyses of children's interactions show that our platform helped children learn good hand hygiene practises. We also found that using a pro-social robot creates enjoyable interactions and greater social engagement between the children and the robot although learning itself wasn't influenced by the pro-sociality of the robot.
\end{abstract}

\section{INTRODUCTION}
Games and social robots captivate and enthrall children. Our study explores the extent to which behaviour change precursors, learning and engagement take place when children and social robots engage in a collaborative game-play around handwashing practises. We first discuss the design and implementation of our novel collaborative educational game titled ``Land of Hands" involving children and a customized social robot (\emph{HakshE}) that we designed. We then present the findings of a user study that we conducted with children aged 6-10 years to evaluate our platform.

Nudge theory postulates that positive and lasting behaviour change such as good handwashing practises could be brought about by repeated nudges that influence the cognitive and affective systems \cite{thaler2009}. Facilitating long-term engagements where teachers and parents nudge children towards good hand hygiene practises could be a tedious process that might take several months to achieve. To address this, we propose to use robots as motivational agents that are designed as playmates. Well-designed robots require minimal human involvement and are not perceptible to challenges such as fatigue. Their novelty, especially in developing countries, attracts children towards them enabling them to be used as agents for nudging  \cite{sreejithIROS}. The long-term goal of this work is both to build an autonomous social robot that promotes positive behaviour change through nudges and to measure the influence of nudges exerted by the robot on children.


\subsection{Collaborative Games for Positive Reinforcement}

Behaviour change may be achieved when children have both declarative knowledge and procedural knowledge about handwashing. The children have to know the steps of handwashing in order to perform it. In our previous research studies, we found that rural and semi-urban school children in India do not have the correct knowledge about hand hygiene steps as prescribed by the World Health Organization (WHO) \cite{amol2019}. Therefore, we decided to blend declarative knowledge and procedural knowledge on hand hygiene into a serious game that the children could play with the social robot. Prior research shows that when robots acted as tutors or peers during gameplay with children, robots were successful in delivering the necessary learning outcomes and encouraged children to remain continuously challenged and motivated \cite{Janssen, Wainer}. 


Bartneck et al. created an opportunity for human participants to engage with a robot through a collaborative version of the game, Mastermind \cite{Bartneck}. They found that when the robot gave unintelligent advice to human players, then the game was perceived as more difficult and the participants rated the value of the robot’s recommendations to be low. When robots acted intelligently and agreeably, they were perceived as being alive and enjoyable. In a similar study, Xin et al. presented a ‘Sheep and wolf’ mixed-reality game, with Sony’s Aibo robots where they reported that the participants considered the robots as teammates and had enjoyable interactions with them \cite{Xin2006}. Thus, we believe that using a social robot to teach good hand hygiene practises to children in a collaborative game-based setting would provide the most enjoyable and positive reinforcement for bringing the behavior change itself.

During the school year 2021-22, many primary schools in India remained closed due to the Covid pandemic. Therefore it was not possible for us to study the influence of our robot's nudges on children's hand washing behaviour change. Instead, as a pilot, we conducted this research study to explore the influence of a social robot's nudges on behaviour precursors, namely children's learning and engagement with the robot. We will be conducting an offline study during the 2022-23 school year as an extension of this work to study the influence on behaviour change.

\subsection{Measuring Social Interactions and Engagement Between Children and Robots}
Literature in the field of HRI states that children engage with a robotic system through various verbal and non-verbal communicative channels such as verbal responses, facial expressions, gestures and gaze \cite{Sidner2003, Sidner2005}. Moreover, a robot's ability to portray pro-social capabilities has been known to positively influence children's engagement with a robot during long-term interactions \cite{muneeb, serholt}. Therefore, to understand how a pro-social robot's nudges can influence children's learning and interactions with the robot in a collaborative game-based setting, we measured children's verbal communication and facial expressions to the robot's responses. 

Through this research study, we aim to answer the following research questions: 
\begin{itemize}
  \item\textbf{Research Question 1}: To what extent does learning about hand hygiene take place in collaborative play between a child and a social robot? 
  \item\textbf{Research Question 2}: To what extent do a pro-social robot's nudges influence the learning, interaction, and engagement of a child with a robot in a collaborative gameplay setting? 
\end{itemize}

\section{\emph{THE CASA PARADIGM} and GAMIFICATION }
The CASA Paradigm which stands for ``Computers As Social Actors" \cite{Nass} along with the Media Equation theory \cite{Reeves} refer to humans treating computers or any software interfaces as fellow humans by displaying characteristic traits such as politeness, assigning gender, etc \cite{Reeves, Johnson}. This behavior is similar to humans' anthropomorphic interpretations of social robots \cite{nicole}. Evidence from prior research supports that the CASA/media equation theory has been increasingly relevant to the domain of HRI \cite{Horstmann, Fischer}.

To facilitate long-term interactions with social robots and improve engagement and retention, elements from the CASA paradigm can be applied to certain gamification principles. This can help in eliminating challenges that the domain of HRI faces such as predictability, lack of excitement and repetitive design of robots \cite{nicole}. Blending gamification principles and the CASA paradigm together can lead to a more enhanced and positive learning outcome for children.

Certain gamification elements such as a rich story narrative, a relevant theme, desirable end-goal, point rewards, and social interaction between the robot and the player set the ground for players to treat social robots as fellow players (social actors) and create continuous engagement opportunities for players \cite{Johnson2016, Johnson2017, Kongeseri1, Kongeseri2}. By playing with one another on the same team to complete a particular task within a given period, players might feel more motivated and encouraged to interact with robots. Including relevant narratives and a storyline that players can relate to, can help increase the player's immersion and engagement in the game \cite{Xi2019}. Having a theme-relevant multi-level game where the robot delivers frequent positive feedback whenever the player successfully progresses to the next task enjoyably promotes further content learning for the players. 
Keeping this in mind we applied certain game elements to the CASA paradigm as discussed above to create a unique educational gaming platform on hand hygiene that helps facilitate learning in children and encourages engagement with a social robot during play.
\begin{figure}[h]
    \vspace{6pt}
    \centering
    \includegraphics[width=\linewidth, height=5.5cm]{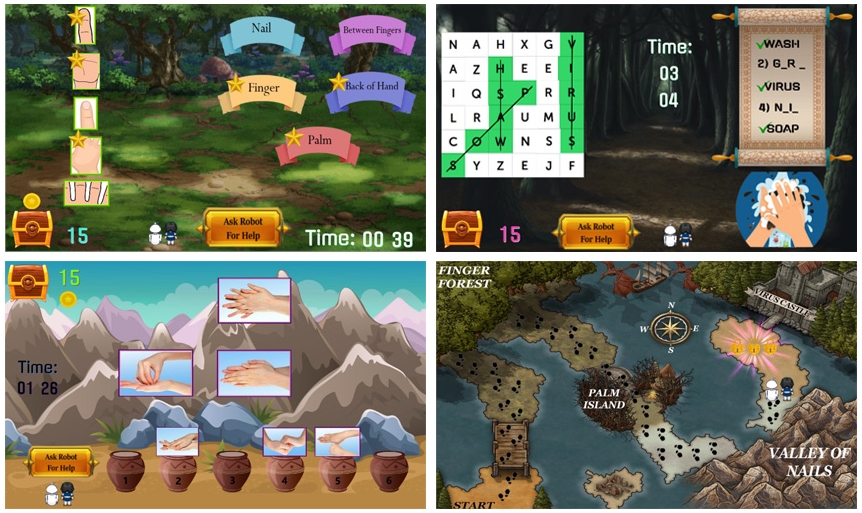}
    \caption{The various levels in the game.}
    \vspace{-10pt}
    \label{fig_levels}
\end{figure}

\section{\emph{LAND OF HANDS}: THE GAME}
\emph{Land of Hands} is a two-player collaborative educational game, based on the theme of hand hygiene. The child and the robot have to play together to rescue a princess who has been captured by germs and locked inside a faraway castle because she did not maintain proper hand hygiene. The game starts with a video narrative that sets the background for the players. The game consists of three levels - A \emph{matching level} that assesses a child's understanding and recognition of various parts of the hand that are involved in the handwashing process; A \emph{word puzzle level} that highlights important vocabulary words related to the topic of hand hygiene; and a \emph{picture ordering} level that helps children remember the correct order of World Health Organisation's (WHO) six steps of handwashing (refer to Figure \ref{fig_levels}). Throughout the game, children interact with \emph{HakshE} to progress to the next level, receive help during the levels, and try to win the game before the allotted time runs out. Upon completing all three levels, the princess gets freed and asks the children to explain to her what they learnt about hand hygiene from their interactions with \emph{HakshE}.
 \begin{figure}[h]
    \centering
    \includegraphics[width=0.94\linewidth]{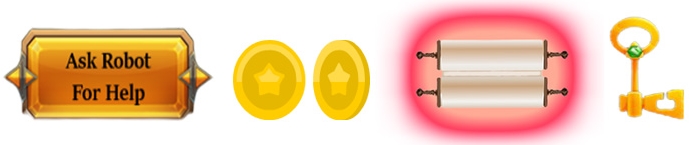}
    \caption{Different game elements such as the help button (social interaction), points (gold coins), scroll, key (rewards).}
    \vspace{-10pt}
    \label{fig_elements}
\end{figure}

The gameplay consists of several elements (refer to Figure \ref{fig_elements}) that have the potential to facilitate a unique engagement in the context of social robotics as a function of people treating robots as social actors [15]. They are:
\begin{itemize}
    \item\textbf{Keys} - There are six golden keys in the game which act as rewards for the player. A key can be won by the child after completing each level and \emph{HakshE} has the other three keys.
    \item\textbf{Scrolls} - There are three glowing scrolls in the game. Each scroll appears before a level begins and allows \emph{HakshE} to ask questions on hand hygiene to the children. If the children answer the question correctly, \emph{HakshE} gives them a key that unlocks the next level.
    \item \textbf{Help Button} - Each of the three levels in the game has an ``Ask Robot For Help" button that allows children to ask \emph{HakshE} for answers or any kind of help while playing. 
    \item\textbf{Points} - Each level has a set of points (gold coins) that the child can earn upon completion of that level. As this is a collaborative game, no points are deducted if the child asks \emph{HakshE} for help.
    \item\textbf{Timer} - Each level has a timer that starts a countdown as soon as the child enters the level. If the level is not completed before the allotted time runs out, the player loses points. To encourage social interactions between \emph{HakshE} and children, the timer is paused whenever they seek \emph{HakshE}'s help.
\end{itemize}
\begin{figure}[h]
    \centering
    \includegraphics[width=0.9\linewidth]{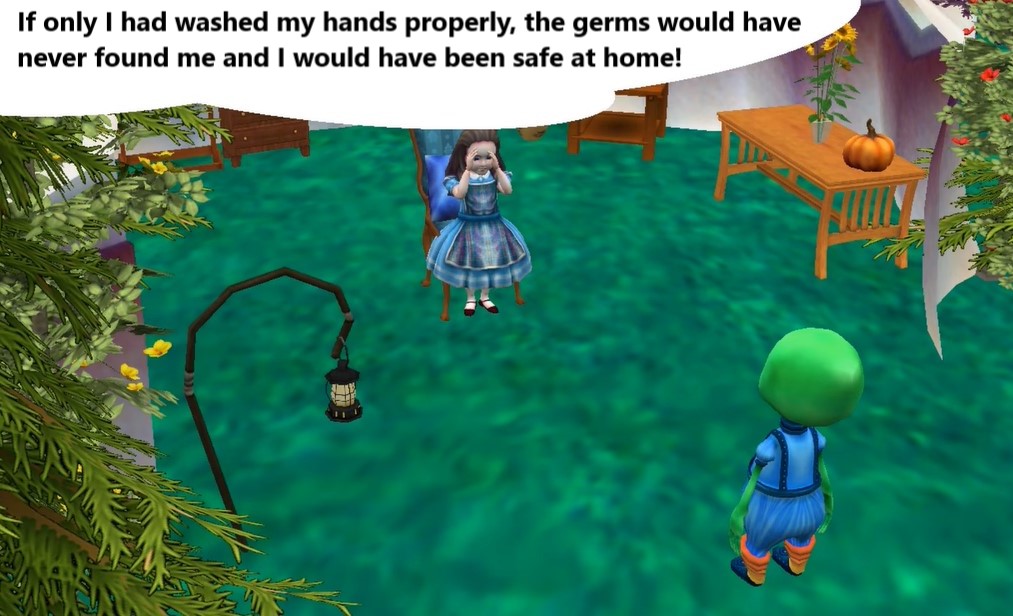}
    \caption{Scene from the video narrative designed using Alice 3 \cite{alice, Cooper}.}
    \vspace{-2pt}
    \label{fig_alice}
\end{figure}

\emph{Land of Hands} is a 2D game that we developed using Godot's game engine (V3.3) \cite{godot}. As shown in Figure \ref{fig_alice}, we used Alice 3 to animate the video narrative that is played at the beginning of the game \cite{alice, Cooper}. The only mode of control required from the player's end is a simple left mouse button click which selects/deselects various elements in the game. The movement of the characters in the game is pre-coded. As this game is meant for young children, we significantly simplified its design to reduce the cognitive load on children. \emph{Land of Hands} can be played in an offline mode with the physical prototype of \emph{HakshE} and in an online mode where both the game and \emph{HakshE} are hosted on \emph{Zoom} video conferencing platform.

\section{METHODS}
\subsection{Study Design}
To explore the influence of a pro-social robot's behavior on learning and interactions between the robot and the child in a collaborative gameplay setting, we split our study into two conditions namely - ``With-Nudges" and ``Without-Nudges". 

In the ``With-Nudges" scenario, the children were provided two kinds of nudges — visual nudges and verbal nudges. Visual nudges were provided in the game in the form of a ``Ask Robot for Help" button that glowed and shook at regular intervals. For the verbal nudge, \emph{HakshE} acted as a pro-social robot exhibiting traits such as proactively offering help. \emph{HaKsh-E} verbally asked the child ``Do you need any help?", when he/she got stuck even if the child did not explicitly ask for the robot's help. \emph{HakshE} also provided constant positive feedback to the child when he/she completed a task. 

In the ``Without-Nudges" scenario, \emph{HakshE} did not display pro-social traits. Unless children specifically called out to \emph{HakshE} for help, the robot did not reach out to them. The levels still contained the ``Ask Robot for Help" button, but it did not glow or shake and was activated only when a child clicked on it.

\subsection{Participants}
A total of 32 children (15 girls and 17 boys) between the ages of 6 to 10 years (M=8.22 years, SD=1.07) were recruited for the study. All the children are students in schools in two different regions of Kerala, India, namely Puthiyakavu and Kochi (Kochi=20 children, Puthiyakavu=12 children). The participants were chosen through convenience sampling to ensure that every child had access to a working laptop and a stable internet connection. As our study consisted of a two-way split, to maintain equality in the distribution of the 32 children, we recruited 16 children for the ``With-Nudges" condition and 16 children for the ``Without-Nudges" condition. The study was undertaken only after the verbal and written consent of the participants’ parents, guardians, and the participants themselves was taken. We ensured that the confidentiality and anonymity of the data collected from the children were maintained throughout the study.

\subsection{Materials}
\subsubsection{HakshE - The Robot}
\begin{figure}[h]
    \vspace{5pt}
    \centering
    \includegraphics[width=0.73\linewidth]{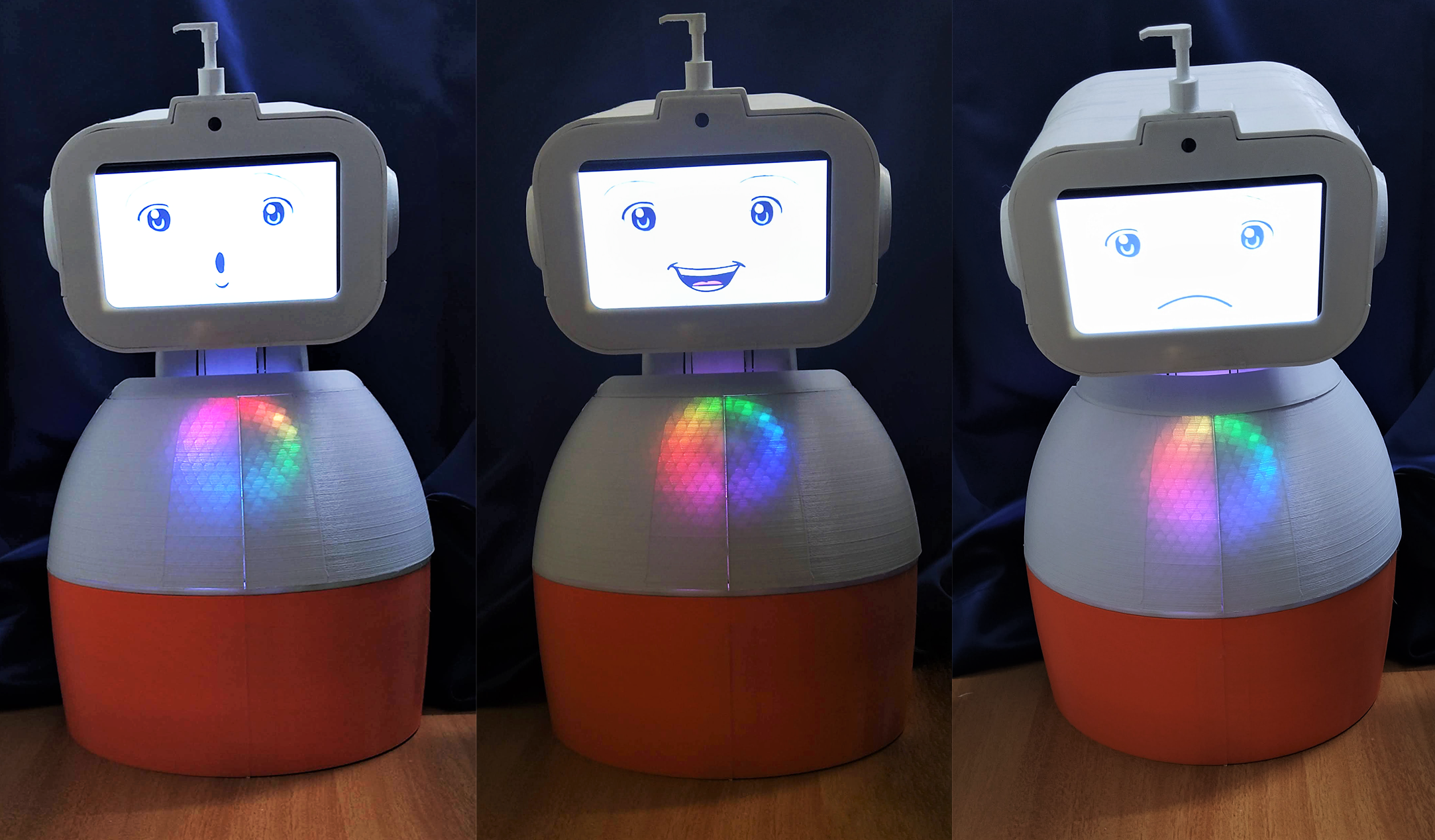}
    \caption{Various emotions displayed by the social robot HakshE during the game.}
    \vspace{-10pt}
    \label{fig_hakshe}
\end{figure}

We custom-designed \emph{HakshE} through a co-design study we conducted with children \cite{sreejithIROS}. \emph{HakshE}'s shell resembles the shape of a soap dispenser to meet its intended requirement of promoting good hand hygiene practises. \emph{HakshE}'s face is an LCD screen that displays a wide range of emotions through its animated facial features. We designed the robot to appear very minimalistic to avoid any unmet expectations that children might have of social robots (refer to Figure \ref{fig_hakshe}). The robot is capable of handling verbal interactions with children on the topic of hand hygiene and essential day-to-day conversations.  

\subsubsection{The Wizard of Oz Platform}
\begin{figure*}[h]
  \vspace{6pt}
  \centering
  \includegraphics[width=0.9\textwidth]{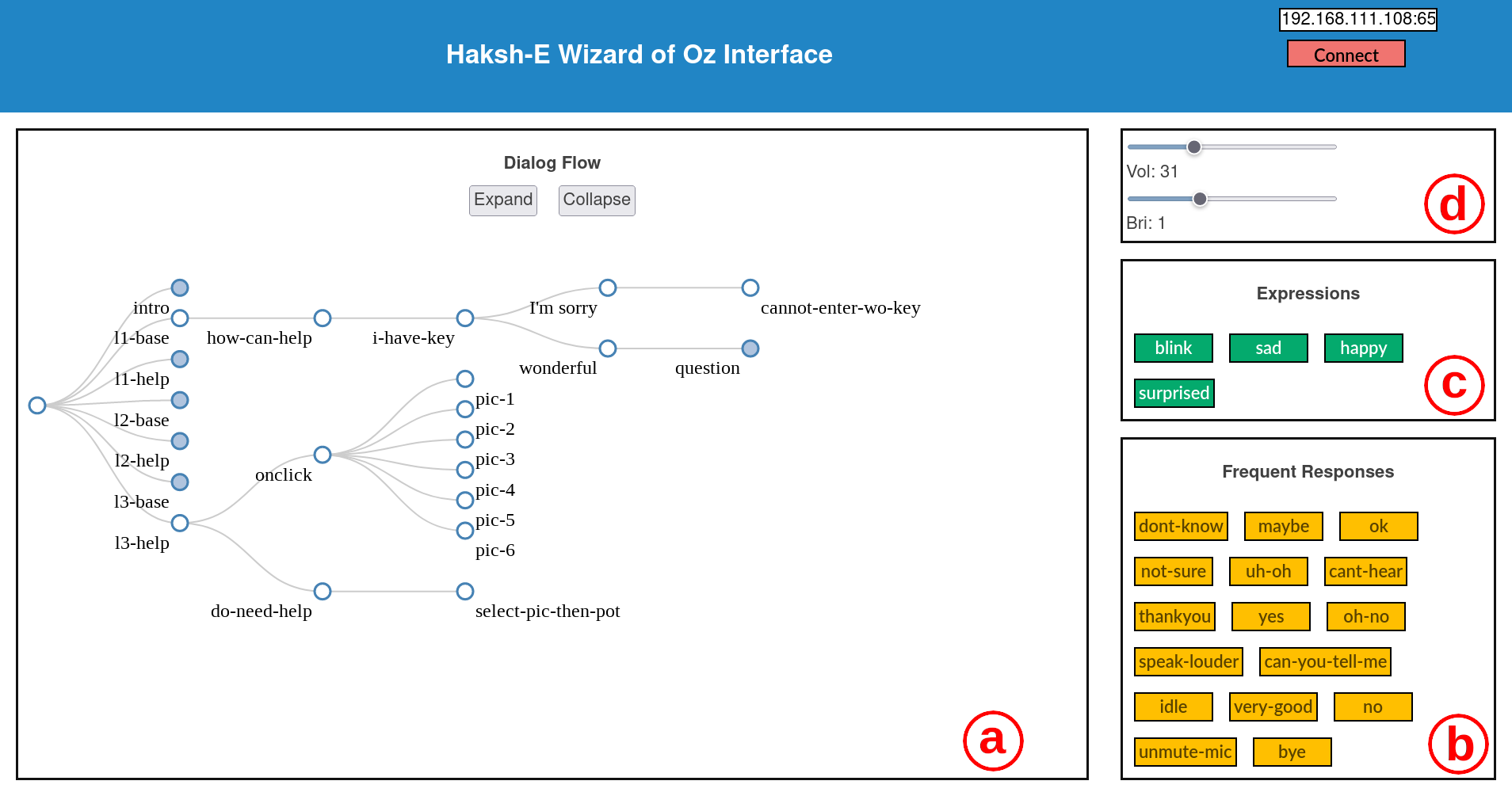}
  \caption{The Wizard of Oz (WoZ) dashboard client. (a) Dialog flow with collapsible branches. Shortened version of dialogues are provided next to each node for ease of identification (b \& c) Frequent responses and expressions of the robot (d) Controls for speech volume and screen brightness.}
  \vspace{-5pt}
  \label{fig_woz}
\end{figure*}

A common criticism against Wizard of Oz \cite{woz} studies in HRI is that such studies can potentially become a Human-Human Interaction scenario \cite{rietz_woz4u_2021,weiss_usus_2009}. This can be tackled by ensuring that the Wizard's interactions are bound to the robot's intended interaction capabilities. Moreover, the interactions of the Wizard should be streamlined to minimize variability across interactions and to reduce the cognitive load for the wizard \cite{rietz_woz4u_2021}. With these two design requirements, we developed a custom Wizard of Oz dashboard for Haksh-E as shown in Figure \ref{fig_woz}. 

We developed the dashboard client to work in any modern web browser, making it platform agnostic. We used \emph{d3.js} to generate an interactive dialog tree with collapsible branches to make it easier for the wizard to keep track of the dialogues. The client communicates with the server via web-socket protocol providing a full-duplex communication channel with minimum latency. The web-socket server, which runs on-board the robot, connects with the robot's API to play the required animation on the LCD screen.  

To avoid inflicting biases on children's behavior after the study, we revealed to them that \emph{HakshE} was teleoperated after the study was conducted. Prior research also shows that in child-robot interaction studies, children's interaction with social robots did not change in any manner when they were informed about the teleoperation post the study \cite{woz1, woz2}.

\subsection{Ethics Statement}
The study was approved by the Clinical Trials Registry (ICMR-NIMS) of India (CTRI/2022/01/039654) and the Institutional Ethics Committee of Amrita Vishwa Vidyapeetham, India (IEC-AIMS-2021-AMC-306). The study was conducted in accordance with the Declaration of Helsinki (DOH). 

\begin{figure}[h]
    \vspace{5pt}
    \centering
    \includegraphics[width=0.8\linewidth, height=4.7cm]{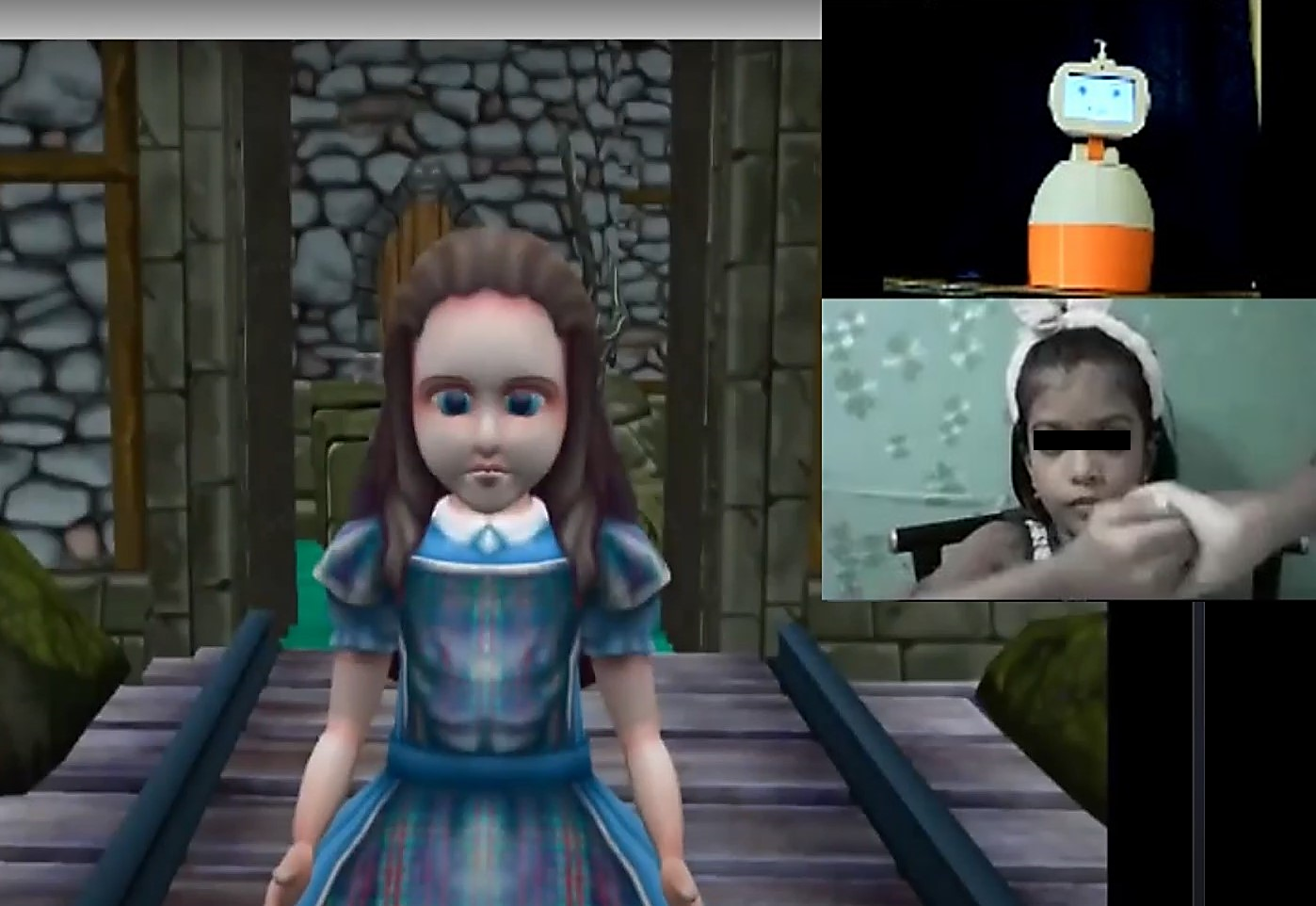}
    \caption{The online setup showing a child demonstrating handwashing steps to the princess.}
    \vspace{-10pt}
    \label{fig:participants}
\end{figure}

\section{USER STUDY}
Due to the ongoing COVID-19 pandemic, we conducted the entire study online through the video teleconferencing platform \emph{Zoom} (refer to Figure \ref{fig:participants}). Children played the game and interacted with the robot individually. Researchers facilitated the video conferencing call to ensure that the data collected is truly a representation of the children’s views. The average time taken to conduct the study with each child was approximately 35 minutes. Since the children did not have prior interaction with a robot, at the beginning of the study, a short general interaction was facilitated between \emph{HakshE} and the child to familiarize him/her with the robot and to see if he/she feels comfortable continuing with the study. One of the researchers explained the rules and instructions of the game to the child. A video on the six steps of handwashing as instructed by WHO was shown to the child before the game. Researchers did not intervene in-between the game unless a technical issue had to be resolved. This allowed children to interact with the robot and play the game on their own, eliminating any kind of bias.
\begin{figure}[h]
    \centering
    \includegraphics[width=0.85\linewidth, height=3.5cm]{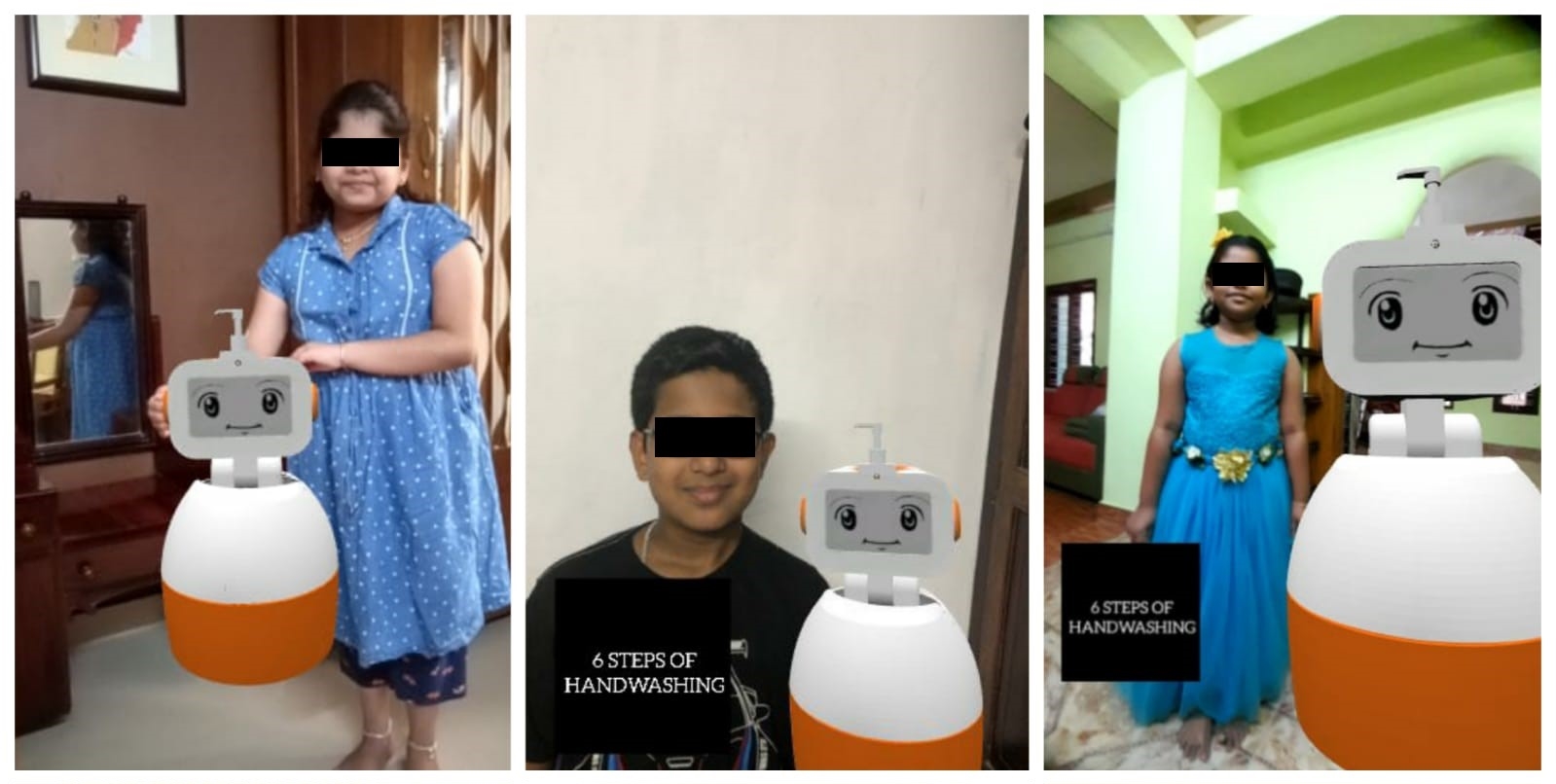}
    \caption{Children's AR experience with HakshE.}
    \vspace{-3pt}
    \label{fig_AR}
\end{figure}

As an incentive for the children, we built a simple web-based Augmented Reality (AR) platform using UniteAR \cite{unitear} where a 3D model of \emph{HakshE} was projected onto their surroundings by scanning a custom QR code using their mobile phones. We conducted a session with the children after the study to explain this AR process. All the children were compensated with a photograph from their AR experience with \emph{HakshE} (refer to Figure \ref{fig_AR}).

\section{RESULTS}
The purpose of our study was to explore the extent of learning about good hand hygiene practises from the gameplay and the influence of a pro-social robot's nudges on the learning and interactions between the child and the robot in a collaborative game-play setting. In this section, we present our findings for our research questions in separate subsections.

\subsection{Learning}
Our first research question was "To what extent does learning about hand hygiene take place in collaborative play between a child and a social robot"? To assess how well children reflected upon what they learnt about hand hygiene through the proposed platform, we devised a rubrics grid \cite{rubrics}. The rubric has three categories: ``Awareness of good and bad hand hygiene habits", ``Knowledge regarding the importance of handwashing", and ``Demonstration/Explanation of the six steps of Handwashing". Evaluation of each category was done on a 3-point scale:
\begin{itemize}
    \item Grade 3 representing ``Independently Answered" - If the child explained what they learnt through the game without any cues from the researcher, 
    \item Grade 2 representing ``Verbal Cues were needed" - If the child required verbal prompts from the researcher to reflect on their learning and 
    \item Grade 1 representing ``Physical Prompts were needed" - If the researcher had to use physical cues to help the child remember what was learnt through the game. 
\end{itemize}

We conducted a two-sample t-test ($\alpha$ = 0.05) to compare the mean rubrics points obtained by the children for the With-Nudges and Without-Nudges conditions. The data was normally distributed and there was homogeneity of variance as assessed by Levene's Test for Equality of Variances. The results showed that there was no statistically significant difference in the points for the With-Nudges (M = 6.4, SD = 0.986) and Without-Nudges (M = 6, SD = 0.756) conditions; t(38) = 1.247, p = 0.223. 

To further understand the extent of learning that took place, we asked the children two questions before and after the study: Question 1 - ``Can you show the princess the handwashing steps you know?" and Question 2 - ``What is the minimum time we should wash our hands for?". 

\begin{figure}[h]
    \vspace{6pt}
    \centering
    \includegraphics[width=0.83\linewidth, height=4cm]{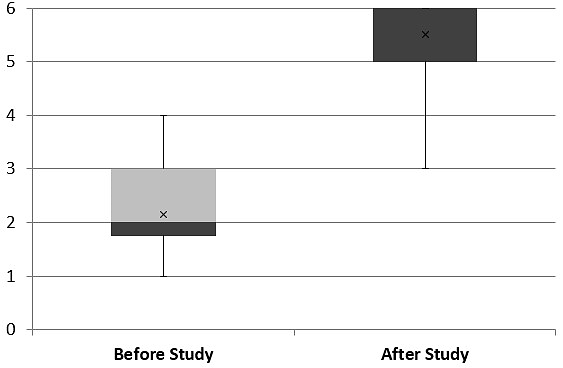}
    \caption{Box plot representing the number of handwashing steps demonstrated by the children before and after the study.}
    \vspace{-5pt}
    \label{fig_box}
\end{figure}

For Question 1, we performed a one-way repeated measures ANOVA ($\alpha$ = 0.05) to compare the mean number of handwashing steps demonstrated by the children before and after the study. The results showed that the number of handwashing steps demonstrated by the children differed statistically significantly before and post the study (F(1, 31) = 407.485, p\textless 0.0005) The box plot in Figure \ref{fig_box} shows the comparison of mean scores pre and post-test. 

For Question 2, according to the Centers for Disease Control and Prevention \cite{cdc}, the minimum duration of handwashing is 20 seconds. Hence, if children answered Question 2 correctly, we coded the data as ``1", and if they were wrong, we coded it as ``0". Since data obtained from Question 2 has a dependent variable that is dichotomous in nature with two mutually exclusive categories (``1" and ``0"), we performed a McNemar's test with correction on the data. The results determined that there was a statistically significant difference in the proportion of correct answers pre-and post-study, p\textless 0.0005. 

We also tested for differences in the learning outcomes between genders but found that there was no statistical significance in any of the above conditions. 

\subsection{Interaction}
Our second research question was "To what extent do a pro-social robot's nudges influence the learning, interaction, and engagement of a child with a robot in a collaborative gameplay setting"? To study the extent of children's social interaction and engagement with \emph{HakshE}, we measured two factors, namely:
\begin{enumerate}
    \item Interaction Level (IL)
    \item Frequency and duration of verbal responses and facial expressions
\end{enumerate}

\textbf{Interaction Level (IL)}: Fridin \cite{Fridin2014} successfully measured the level of interaction (IL) between children and robots using a unique index based upon the the \emph{KindSAR} interaction measurement index \cite{Fridin2011}. From their study, we adopted the following index to measure the quality of child-robot interaction at a stage ``S'' in the interactive session:\\

$IL_s$ = $EC_s$ * $Sign_s$ * $ \sum_{F=1}^{3} W_FF$\\

Where, $IL_s$ = Interaction level, $EC_s$ = Eye contact with the proposed platform, $Sign_s$ = Variable that indicates positive or negative interaction ($Sign_s$ = +1 if child has a positive interaction with the platform, $Sign_s$ = -1 if child has a negative interaction with the platform; measured by a two-dimensional valence and arousal space \cite{ueki1994}), F = Affective Factor and $W_F$ = Binary variable ($W_F$ = 1 if child expresses emotion, $W_F$ = 0 if child does not express any emotion).

Here, the variable EC=3 if the child looks at the platform during the interaction, EC = 1 if the child looks at the researcher for help/explanation during the interaction, and EC=0 if the child does not interact with the platform at all. We consider two affective factors in our study - facial expressions and verbal responses. The variable F=1 if the child expresses facial emotions (such as smiles and surprised reactions) and F=3 if the child expresses emotions via verbal responses. 

We acknowledge that our study has limitations due to its online nature. We could not control the positioning of the children in front of their laptop cameras. This led to the video only capturing children's faces up-till their neck most of the time and not their entire bodies. Hence other social engagement cues such as gestures could not be captured and measured.

We performed detailed video analysis of the data and manually calculated the values of IL. In total there were 797 interactions from the study population (n=32). We further categorized the IL data according to the Nudges conditions (With-Nudges Vs Without-Nudges) and the gender conditions (Boys Vs Girls). Inspection of the Shapiro-Wilk test revealed that the IL data was non-normally distributed for both the nudges conditions and the gender conditions. Also, there was no homogeneity of variance as assessed by Levene’s Test for Equality of Variances. Hence, we performed a robust Welch ANOVA followed by a Games-Howell post-hoc test on our data.

Welch's ANOVA determined that the mean IL scores differed statistically significantly between the With-Nudges and Without-Nudges conditions (F (1,797) = 12.721; p = 0.00038). The Games-Howell post-hoc test also revealed statistical significance between the independent groups (p = 0.00038; the mean of the With-Nudges condition was greater than the mean of the Without-Nudges condition). 
Similarly, IL was also significantly affected by gender (F (1,797) = 14.163; p = 0.00018), with the means significantly higher for girls than boys.

\textbf{Frequency and duration of verbal responses and facial expressions}: Next, we measured both the frequency and duration (seconds) for both the nudges conditions. We used ELAN software \cite{elan} to transcribe and categorize each utterance and expression made by the children into two factors - ``Verbal Responses" and ``Facial Expressions". 
\begin{figure}[h]
    \centering
    \subfloat[\centering Frequency]{{\includegraphics[width=0.82\linewidth]{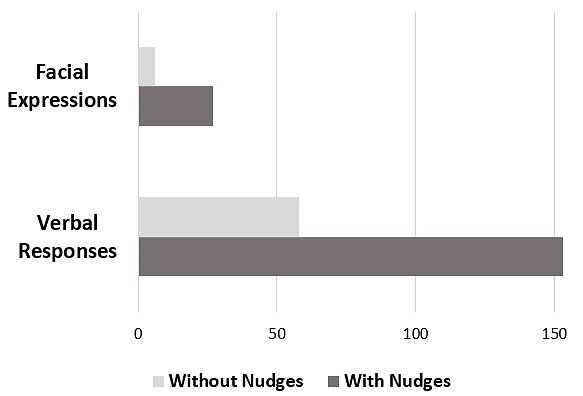} }}
    \qquad
    \subfloat[\centering Duration]{{\includegraphics[width=0.82\linewidth]{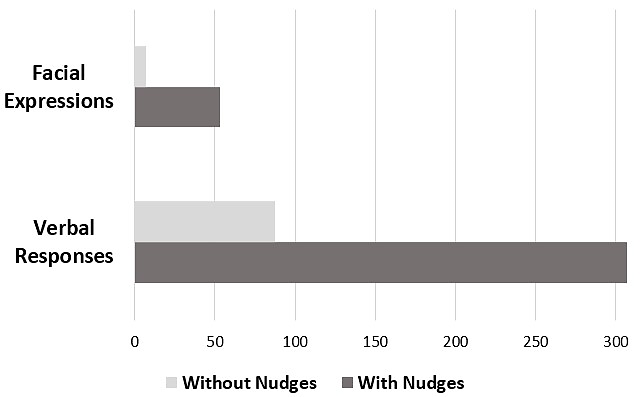} }}
    \caption{The frequency and duration of both the verbal responses as well as the facial expressions are more in the pro-social robot scenario.}
    \vspace{-10pt}
    \label{fig_bar}
\end{figure}

\begin{itemize}
    \item \textbf{Verbal Responses}: Children's responses to the robot such as words, sentences, and vocalisations were considered to be verbal responses \cite{serholt}. We observed that the most common verbal responses made by the children during the gameplay were: ``Help me", ``Okay", ``Thank you", ``Yes" and ``No". Utterances such as ``What" and ``Huh" were not considered to be a sign of social engagement with the robot as it generally depicted situations where children could not understand or hear what the robot was saying. Within a 5-second time frame, if there was a change in the context of what the child was saying, we counted it as two different responses.
    \item \textbf{Facial Expressions}: Children's facial expressions that signify positive social engagement with a robot include ``Smiling" and ``Surprised" expressions \cite{muneeb, serholt}. All types of smiles and surprised expressions of the children in response to the robot were considered to be positive facial expressions. Expressions such as ``Frustration" and ``Grinding teeth" were not considered to be signs of social engagement. The duration of children's smiles and surprised reactions were calculated until we observed a change in their facial expressions.
\end{itemize}

Inspection of Q-Q Plots revealed that the frequency and duration data for both conditions were not normally distributed. Therefore, we conducted a non-parametric Mann–Whitney U test ($\alpha$ = 0.05) for these two factors in the With-Nudges and Without-Nudges conditions. The results (refer to Figure \ref{fig_bar}) are summarized below.
\begin{itemize}
    \item \emph{Frequency of Verbal Responses}: Frequency of verbal responses in the With-nudges condition (Mdn = 6.5) was higher than that of the Without-nudges condition (Mdn = 2.5). The Mann–Whitney U test indicated that this difference was statistically significant, U(N$_{With-Nudges}$ = 16, N$_{Without-Nudges}$ = 16) = 71.5, z = 2.119, p = 0.031.
    \item \emph{Duration of Verbal Responses}: Duration of facial expressions in the With-Nudges condition (Mdn = 6.32) was higher than that of the Without-Nudges condition (Mdn = 3.798). The Mann–Whitney U test indicated that there was no statistically significant difference, U(N$_{With-Nudges}$ = 16, N$_{Without-Nudges}$ = 16) = 91, z = 1.38, p = 0.171.

    \item \emph{Frequency of Facial Expressions}: Frequency of facial expressions in the With-Nudges condition (Mdn = 1) was higher than that of the Without-Nudges condition (Mdn = 0). The Mann–Whitney U test indicated that this difference was statistically significant, U(N$_{With-Nudges}$ = 16, N$_{Without-Nudges}$ = 16) = 65, z = 2.721, p = 0.016.
    \item \emph{Duration of Facial Expressions}: Duration of facial expressions in the With-Nudges condition (Mdn = 2.075) was higher than that of the Without-Nudges condition (Mdn = 0). The Mann–Whitney U test indicated that this difference was statistically significant, U(N$_{With-Nudges}$ = 16, N$_{Without-Nudges}$ = 16) = 61, z = 2.882, p = 0.01.
\end{itemize}

We once again tested for differences in interaction and engagement between genders but found that there was no statistical significance in any of the above conditions. 

\section{DISCUSSION}

We identify two major findings that answer our two research questions, which we present below:

\textbf{Extent of Learning from the Proposed Platform}: Our analyses indicate that children's learning and retention from the proposed platform are significantly higher post-study when compared to pre-study data. This tells us that our platform was successful in educating children on hand hygiene. Interestingly we observed that children's learning was not influenced by the pro-sociality of the robot. Also, both boys and girls learnt equally well.

\textbf{Influence on Children's Engagement and Interactions with the Robot}: From our results, we conclude that the overall Interaction Level (IL) between the children and the robot was significantly higher in the pro-social scenario. Also, based on the IL index, it is interesting to note that girls interacted more with \emph{HakshE} compared to boys. The frequency and duration of facial expressions and the frequency of verbal responses are significantly higher in the pro-social robot scenario than in the non-pro-social robot scenario. Though the duration of verbal responses in both the nudges conditions did not differ statistically significantly, from Figure \ref{fig_bar}(b), it is evident that the children verbally interacted with the robot more in the With-Nudges condition, even if they were not very verbose when they interacted with the robot. Thus we can say that a pro-social robot in a collaborative game-play setting increases positive social engagement with children. Also, the results tell us that gender differences in the above conditions do not influence children's interaction with the robot in the pro-social vs non-pro-social case.

\section{CONCLUSION}
In this paper, we set out to explore whether an educational collaborative game-play platform with a pro-social robot (\emph{HakshE}) can help children learn about good hand hygiene practises. We also explore the influence of the platform on children's engagement and interactions with the robot. We start by proposing the design of our game - ``Land of Hands" based on the theme of hand hygiene using Godot's gaming engine and Alice 3. We then present the findings of a user study that we conducted with children aged 6-10 years to evaluate the effectiveness of our platform. We split our study into two conditions - ``With-Nudges" and ``Without-Nudges" to understand the influence of a pro-social robot's nudges on children's learning and interaction. We conclude that a collaborative gameplay scenario with a pro-social robot and children is successful in not only delivering the required learning outcomes but also in creating enjoyable interactions and engagement. 

In the 2022-23 school year, we intend to conduct an in-person study using the proposed platform. We will explore the extent to which a pro-social robot affects other social engagement cues such as gestures and the gaze of the children. We also intend to evaluate the model in a long-term child-robot scenario to study the effect of the gameplay with the social robot on behaviour change itself.

\section{ACKNOWLEDGMENTS}
We would like to express our gratitude to Sri Mata Amritanandamayi Devi (Chancellor), a world-renowned humanitarian and spiritual leader without whose guidance and constant encouragement, this project would not have been possible. We would like to thank AMMACHI labs for providing seed funding for the project. Finally, we would like to thank Dr. Bhavani Rao and Akshay Nagarajan for their continual support and valuable input during our study.

\bibliographystyle{IEEEtran} 
\bibliography{IEEEabrv, IEEEexample}

\addtolength{\textheight}{-12cm}   

\end{document}